\pdfoutput=1

\documentclass[11pt]{article}

\usepackage{ACL2023}

\usepackage{graphicx}
\graphicspath{ {./images/} }

\usepackage{times}
\usepackage{latexsym}
\usepackage{amssymb}
\usepackage{amsmath}
\usepackage{bbm}
\usepackage{multirow}

\usepackage[T1]{fontenc}

\usepackage[utf8]{inputenc}

\usepackage{microtype}

\usepackage{inconsolata}
\newtheorem{remark}{Remark}[section]

\title{Now It Sounds Like You: Learning Personalized Vocabulary On Device}

\author{Sid Wang$^*$ \\ Meta Reality Labs \\
  \texttt{yuwang2020@meta.com} \And
  Ashish Shenoy$^*$ \\ Meta Reality Labs \\
  \texttt{ashishvs@meta.com} \AND
  Pierce Chuang \\ Meta Reality Labs \\
  \texttt{pichuang@meta.com}  \And
  John Nguyen \\ FAIR, Meta \\
  \texttt{ngjhn@meta.com} 
 }

\begin{document}
\maketitle
\def\thefootnote{*}\footnotetext{Equal contribution.}
\begin{abstract}

In recent years, Federated Learning (FL) has shown significant advancements in its ability to perform various natural language processing (NLP) tasks. This work focuses on applying personalized FL for on-device language modeling. Due to limitations of memory and latency, these models cannot support the complexity of sub-word tokenization or beam search decoding, resulting in the decision to deploy a closed-vocabulary language model. However, closed-vocabulary models are unable to handle out-of-vocabulary (OOV) words belonging to specific users. To address this issue, We propose a novel technique called "OOV expansion" that improves OOV coverage and increases model accuracy while minimizing the impact on memory and latency. This method introduces a personalized "OOV adapter" that effectively transfers knowledge from a central model and learns word embedding for personalized vocabulary. OOV expansion significantly outperforms standard FL personalization methods on a set of common FL benchmarks. 

\end{abstract}

\section{Introduction}

\begin{figure*}[t]
     \centering
     \begin{minipage}[t]{0.4\linewidth}
         \centering
         \includegraphics[width=\linewidth]{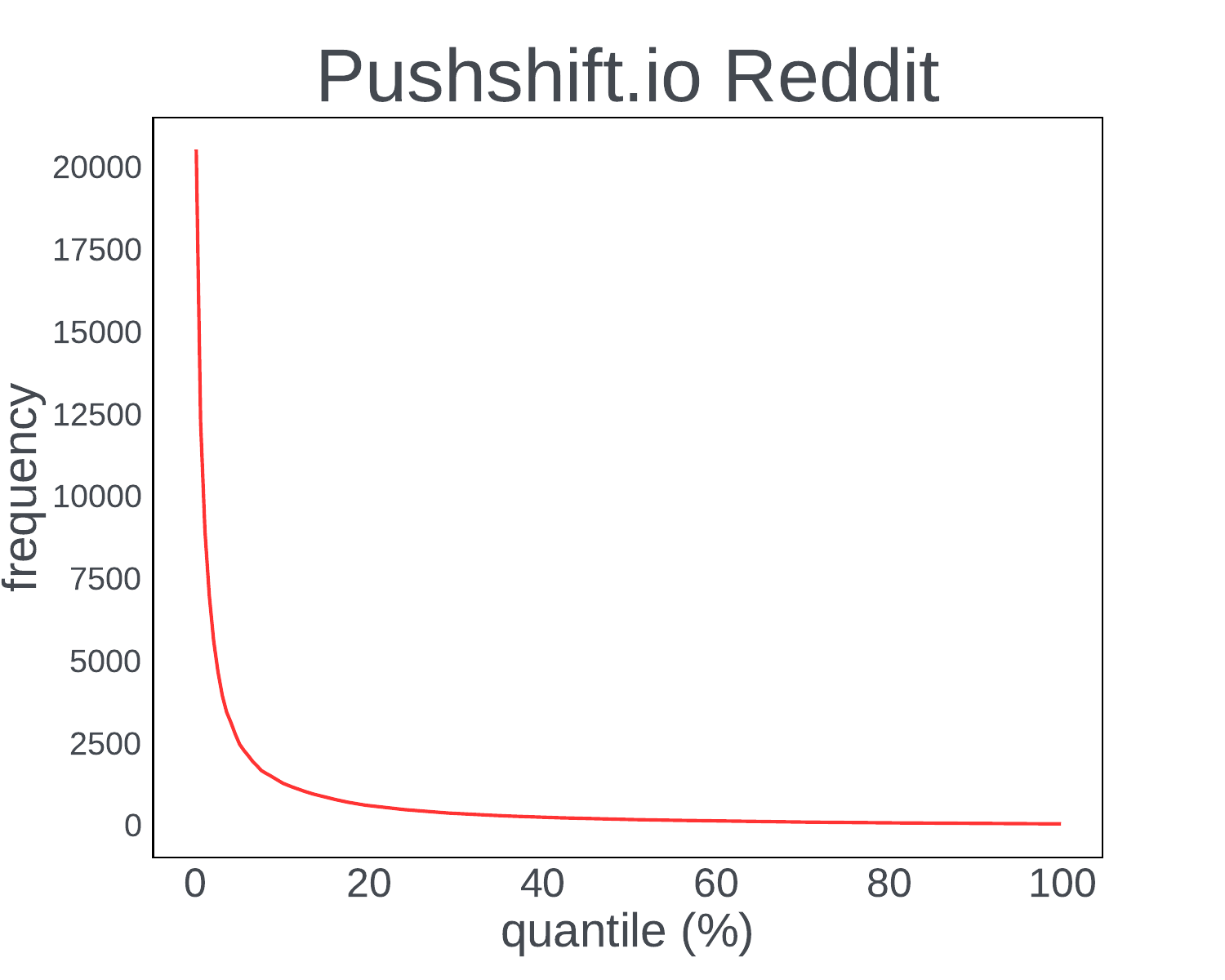}
     \end{minipage}
     \begin{minipage}[t]{0.4\linewidth}
         \centering
         \includegraphics[width=\linewidth]{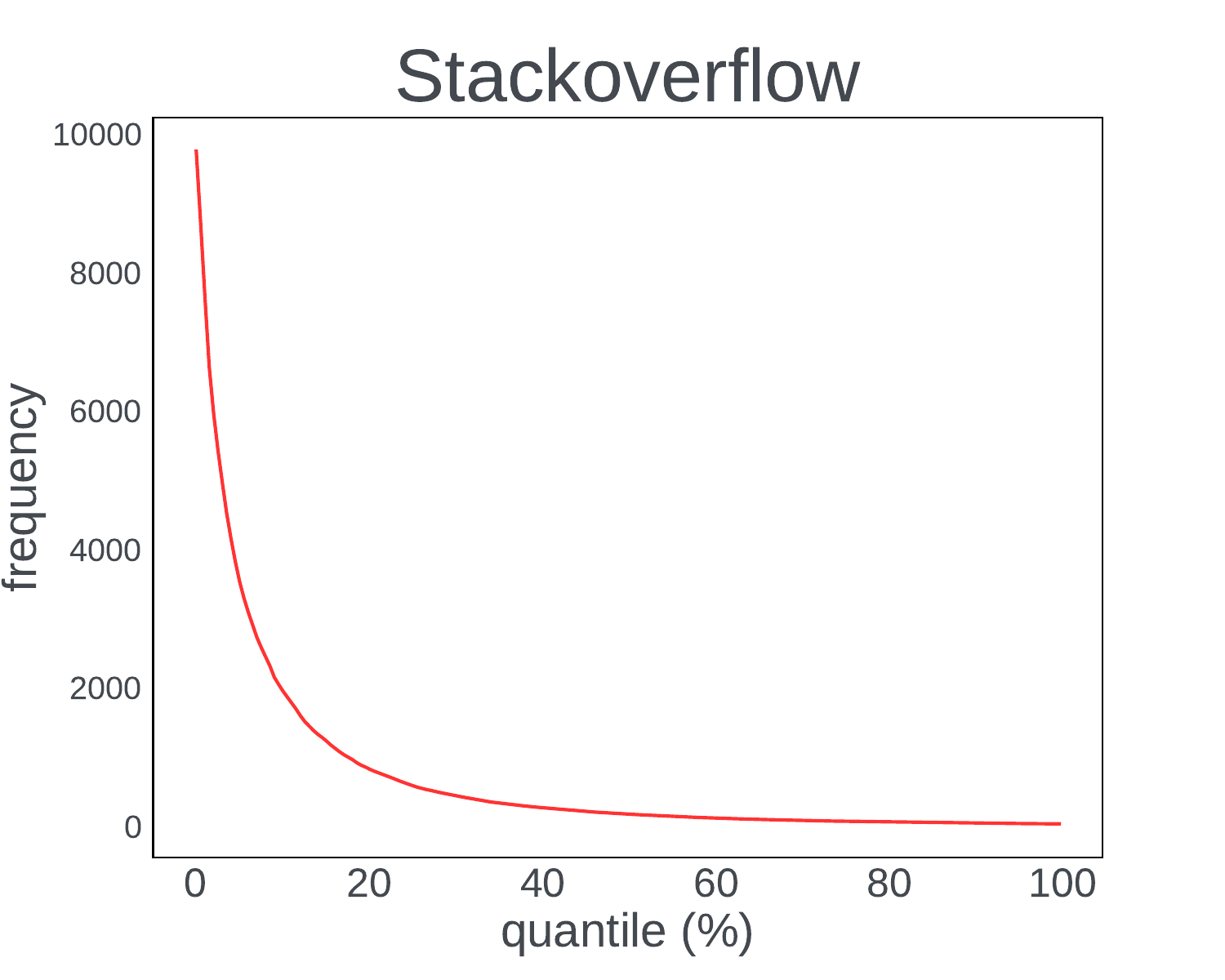}
     \end{minipage}
     \caption{The quantile plot of word frequency for top 10k words from 2 datasets that demonstrates the long-tail phenomenon of OOV.}
     \label{oov_tail}
\end{figure*}

Federated learning (FL) is a distributed machine learning paradigm that enables model training on decentralized datasets without exchanging or sharing sensitive data \cite{McMahan2016CommunicationEfficientLO}. Personalized FL considers models unique to each client under heterogeneous data settings \cite{Hanzely2020FederatedLO}. During personalization, the server sends a global model to the clients for local fine-tuning then the client uses that model for inference. Our work studies a class of on-device language models for next-word prediction task trained with personalized FL.

The resource-constrained edge devices \cite{Chen2019GmailSC,Qiu2022ZeroFLEO,Mathur2021OndeviceFL,yousefpour2023green} limits the usage of subword-level tokenizers. On the one hand, subword tokenizers with large vocabulary require a large memory footprint, making deployment infeasible. On the other hand, a smaller vocabulary leads to longer tokenized sequences and increases generation latency. In order to satisfy these constraints, a \emph{closed vocabulary} \cite{Qin2013FindingRO}, a word-level vocabulary with a white-space tokenizer that treats all unknown words as a special token, must be used. 

Consequently, the model cannot handle out-of-vocabulary (OOV) words \cite{Chen2019FederatedLO}, making it more challenging to understand the communication style of an individual user. Previous methods have demonstrated effectiveness of user specific adaptation using domain embeddings \cite{ashish_interspeech} or prompt tuning \cite{saket_wecnlp}. But
the heavy tail of OOV (shown in Figure \ref{oov_tail}) motivates the need for personalized vocabulary which they do not address. In this work, we propose \emph{OOV Expansion} to personalize on-device vocabulary, tailoring the model toward users' unique wording habits and spelling patterns. \emph{OOV Expansion} uses an adapter -- an MLP with residual connections inserted to different submodules to help the model adapt to new knowledge domain \cite{Houlsby2019ParameterEfficientTL} -- to compute the embedding output of the OOV words. \\
\\
\textbf{Our contributions} We highlight the main contributions:
\smallskip

    $\bullet$ We propose \emph{OOV Expansion}, a novel personalized FL method, to extract useful features for user-specific vocabulary and address the long tail OOV issue.
    
    $\bullet$ We perform comprehensive experiments, showing superior results over that of the baselines, across a set of standard FL datasets.
    
    $\bullet$ We demonstrate that our technique significantly outperforms previous methods with up to $5.6\%$ relative improvements on next-word prediction accuracy and reducing the averaged unknown-word-rate by at least $97\%$.

\section{Method}

This section describes the model architecture and outlines the training techniques for the on-device next-word prediction task.

\begin{figure}[h]
\centering
\includegraphics[width=0.45\textwidth, height=12cm]{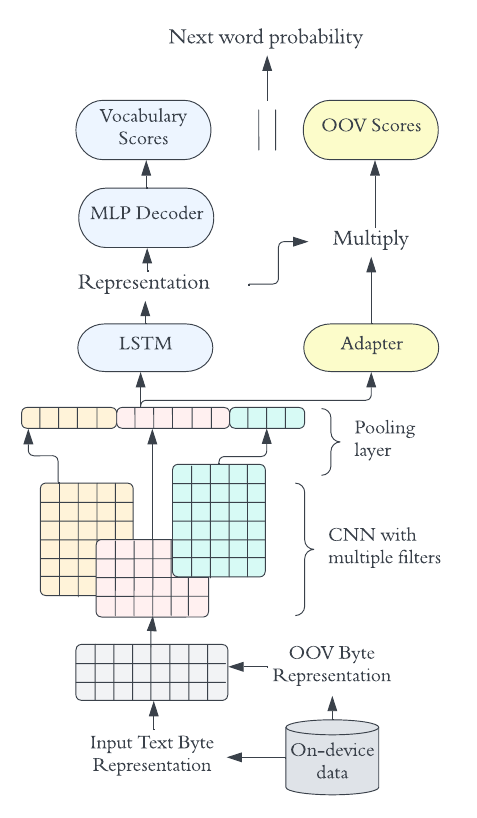}
\caption{The mechanism flowchart of the OOV Expansion stage; here $\|$ denotes vector concatenation.}
\label{oov_design}
\end{figure}
\subsection{Character-aware language model}
\label{origmodel}

The limitations on the memory and latency of on-device modeling confines the model architecture and rules out some common choices such as the transformers \cite{Vaswani2017AttentionIA}. Instead, we use the character-aware LSTM based language model, a commonly adopted architecture for on-device applications \cite{Kim2015CharacterAwareNL,Hard2018FederatedLF}. 

The model consists of 3 major components: (1) a character level CNN that computes word embeddings, which will be denoted as CharCNN from now on, (2) an LSTM encoder that provides the input representations, (3) an MLP decoder that gives the scores over vocabulary. See the non-yellow stack in Figure \ref{oov_design}.

\begin{remark}
\label{score}
The term ``score'' is an important and recurring notion throughout the section. The scores of multiple sources of vocabulary may be joined and passed through a LogSoftmax transform to produce logits over the vocabulary union, which are eventually used to compute the cross entropy loss for next-word prediction.
\end{remark}

The model hyperparameters are defined as follows: the CharCNN contains a utf-8 embedding of shape $\mathbb{R}^{256 \times E}$ and a CNN with $D$ channels and kernel size $K$. The LSTM encoder contains $M$ layers with both input and output dimensions equal to $D$. The decoder is a linear layer that has input dimension $D$ and output dimension $|\mathcal{V}|$, with $\mathcal{V}$ being the closed vocabulary.

\subsection{Baseline 1: OOV-as-UNK}
\label{baseline1}
This is the traditional FL personalization (i.e. involving only the non-yellow stack in Figure \ref{oov_design}). To be precise, the model is first pretrained on a general dataset using the standard “next-word prediction” task, and then sent to client devices for federated learning. Lastly, perform personalization without any additional treatments for OOVs rather than replacing them with the special token [UNK]. 

\subsection{Baseline 2: OOV-oracle}
\label{baseline2}
The second baseline assumes to have the knowledge of all users' OOV on server. However, because of the tight memory budgets in practice we cannot afford full vocabulary training, we instead expand the OOV-as-UNK vocabulary by the $N$ most frequent OOV words. Upon expansion, the rest stages (including pretraining, FL, and personalization) proceed in exactly the same way as in Subsection \ref{baseline1}.

\subsection{Our method: OOV Expansion}
\label{oovmodel}

We start with the unpersonalized OOV-as-UNK (ref. Subsection \ref{baseline1}) and perform personalization as demonstrated in Figure \ref{oov_design}. To begin, given an input sentence the model forwards in exactly the same fashion as in subsection \ref{origmodel} to get the vocabulary scores (ref. Remark \ref{score}). Next, retrieve the client's top $n$ OOV words and compute their representations from CharCNN, and further pass them through an adapter (i.e. a residual MLP that is randomly initialized at the beginning of personalization on each device) with hidden dimensions $\vec{\textbf{H}} = (H_1, H_2, \cdots, H_L)$ where $L$ is the number of layers in the adapter. This intermediate result could be interpreted as “adapted OOV word embeddings”. Then take the inner product between the adapter outputs and the LSTM encoder outputs to get OOV scores (ref. Remark \ref{score}). Finally, concatenating the OOV scores with the vocabulary scores, we obtain the full scores over the client personalized vocabulary for next word prediction.

To understand the functionality of adapter, note that the model never learns to compute OOV representations during pretraining or federated learning stage because unknown words are ignored and replaced by a special token. During personalization, the CharCNN module mutlitasks on (1) providing inputs for LSTM encoder and (2) computing OOV embeddings. Morally speakng, OOV adapter alleviates the interference between the two tasks as well as adapting the prior knowledge of CharCNN to the new “OOV task”.

Lastly, our design guarantees that no sensitive OOV information ever leaves the client device and hence mitigates the privacy leakage risk while preserving the model's ability to learn OOV features.

\begin{figure*}[t]
     \centering
     \begin{minipage}[t]{0.4\linewidth}
         \centering
         \includegraphics[width=\linewidth]{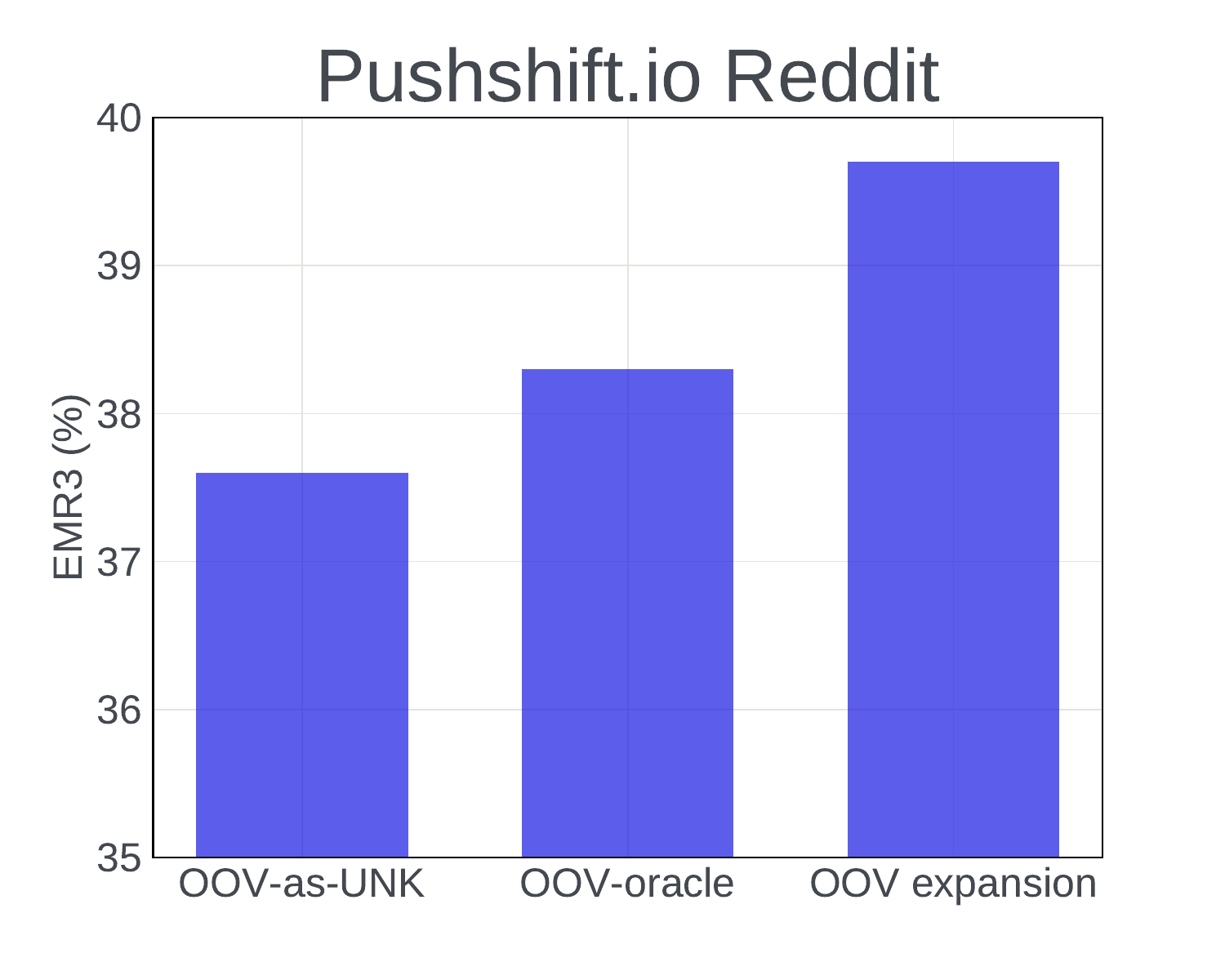}
     \end{minipage}
     \begin{minipage}[t]{0.4\linewidth}
         \centering
         \includegraphics[width=\linewidth]{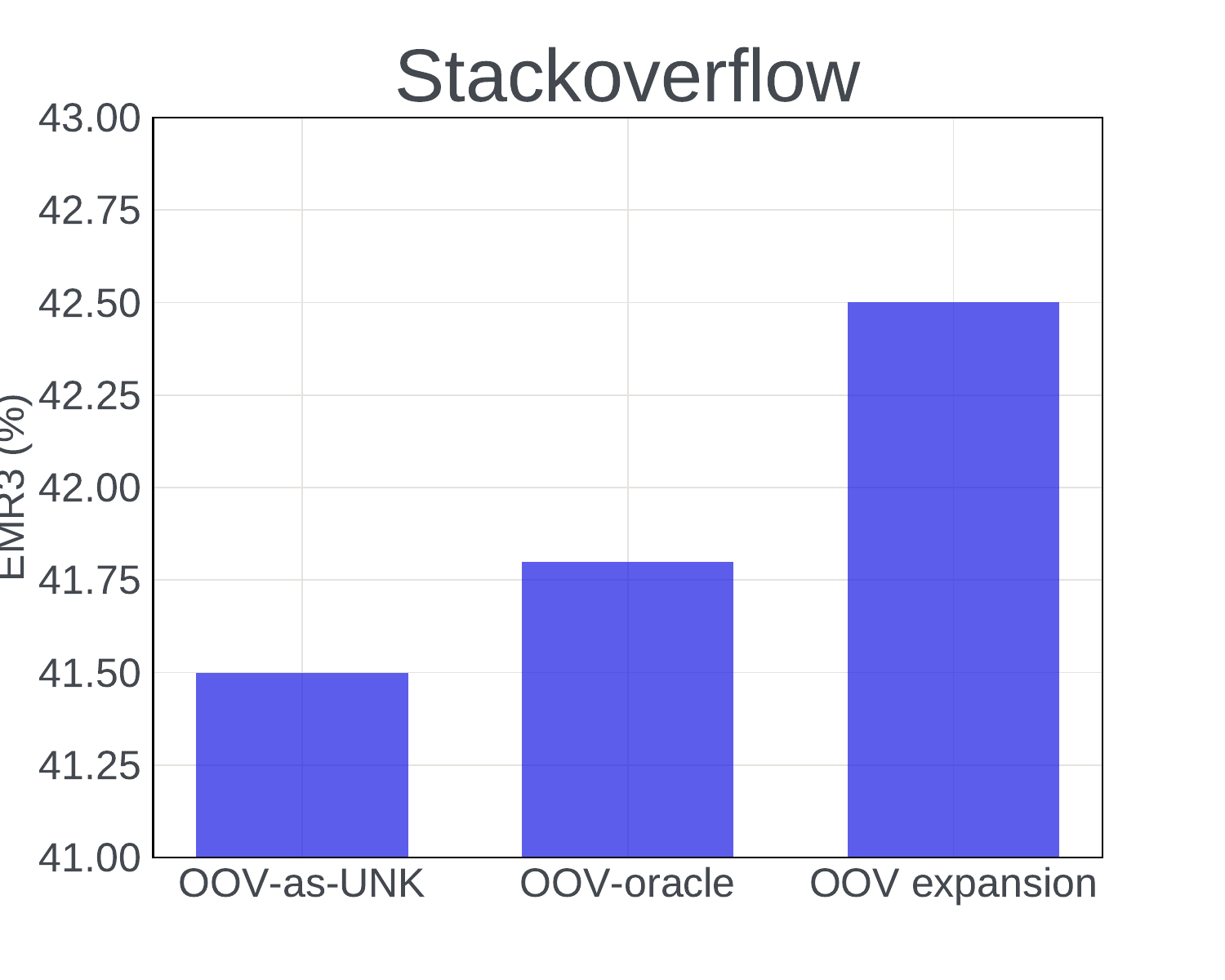}
     \end{minipage}
     \caption{$\text{EMR}_3$ of OOV Expansion and two baselines on 2 datasets.}
     \label{main_results}
\end{figure*}

\section{Experiments}

\subsection{Model setup}

The model hyperparameters in Subsection \ref{origmodel} are chosen as $E=100$, $D=200$, $K=4$, $M=2$. 

To initialize OOV-as-UNK (ref. Subsection \ref{baseline1}) we extract the 5k most frequent words from a centralized FL dataset (see subsection \ref{fldata}) to form a closed vocabulary, and pretrain the model on wikitext-103 \cite{Merity2016PointerSM} using next-word prediction task for 100 epochs with learning rate as $10^{-4}$, batch size equals to 32, across 8 GPU devices, and evaluated by the exact match metric $\text{EMR}_3$ (see Subsection \ref{evalmetric} for the precise definition). The OOV-oracle baseline (ref. Subsection \ref{baseline2}) follows the same configuration as OOV-as-UNK, except for having a 10k-sized vocabulary. In other words, OOV-oracle expands the vocabulary of OOV-as-UNK by $N=5000$ most frequent OOVs. The OOV-as-UNK (resp. OOV-Oracle) model has 1.76M (resp. 2.76M) parameters in total, and took up to 36 hours on 8 Nvidia V100 16GB GPUs to complete pretraining. 

During OOV expansion (see Subsection \ref{oovmodel}), we personalize up to 1000 most frequent out-of-vocabulary words from each decentralized dataset.

\begin{figure*}[t]
     \centering
     \begin{minipage}[t]{0.4\linewidth}
         \centering
         \includegraphics[width=\linewidth]{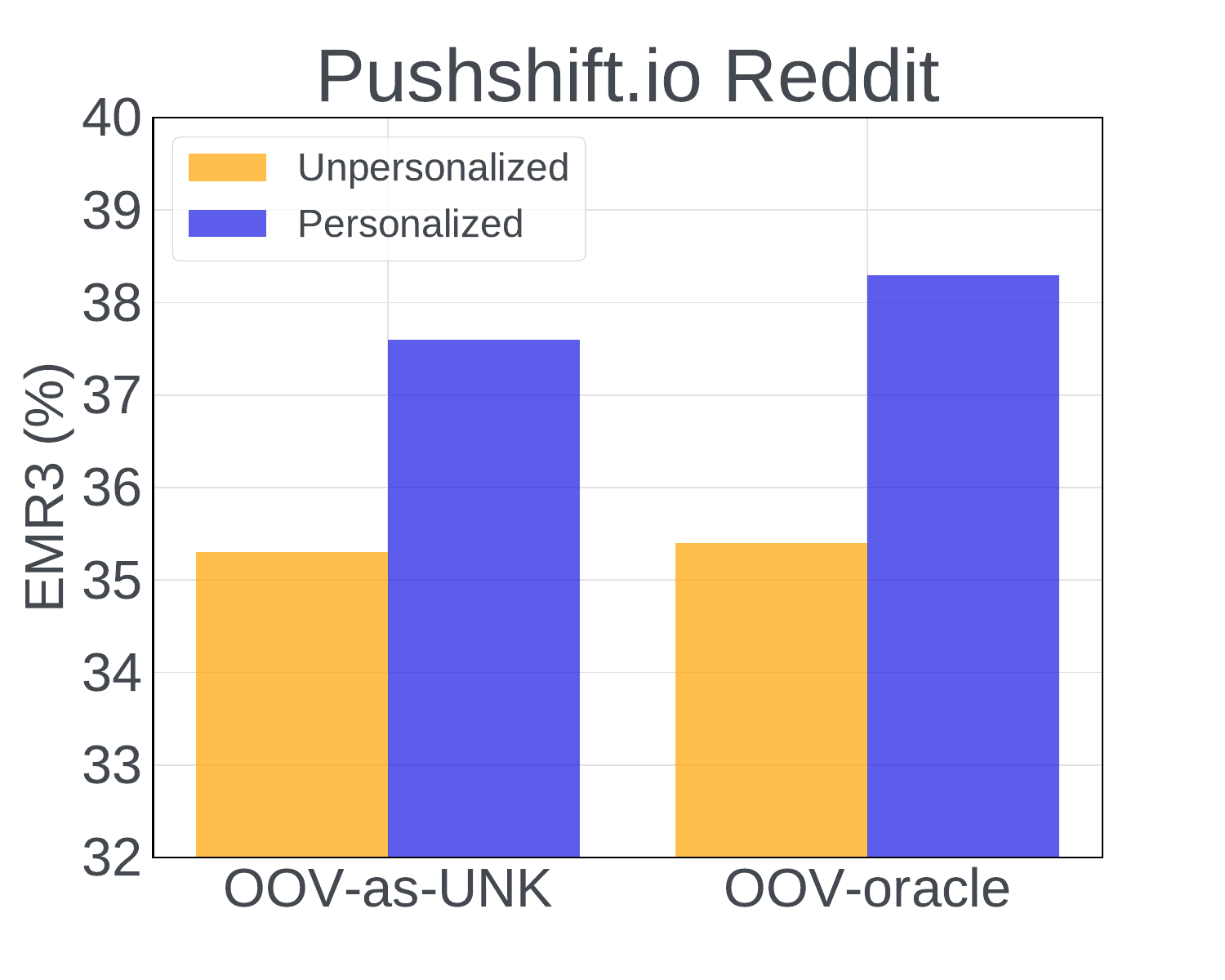}
     \end{minipage}
     \begin{minipage}[t]{0.4\linewidth}
         \centering
         \includegraphics[width=\linewidth]{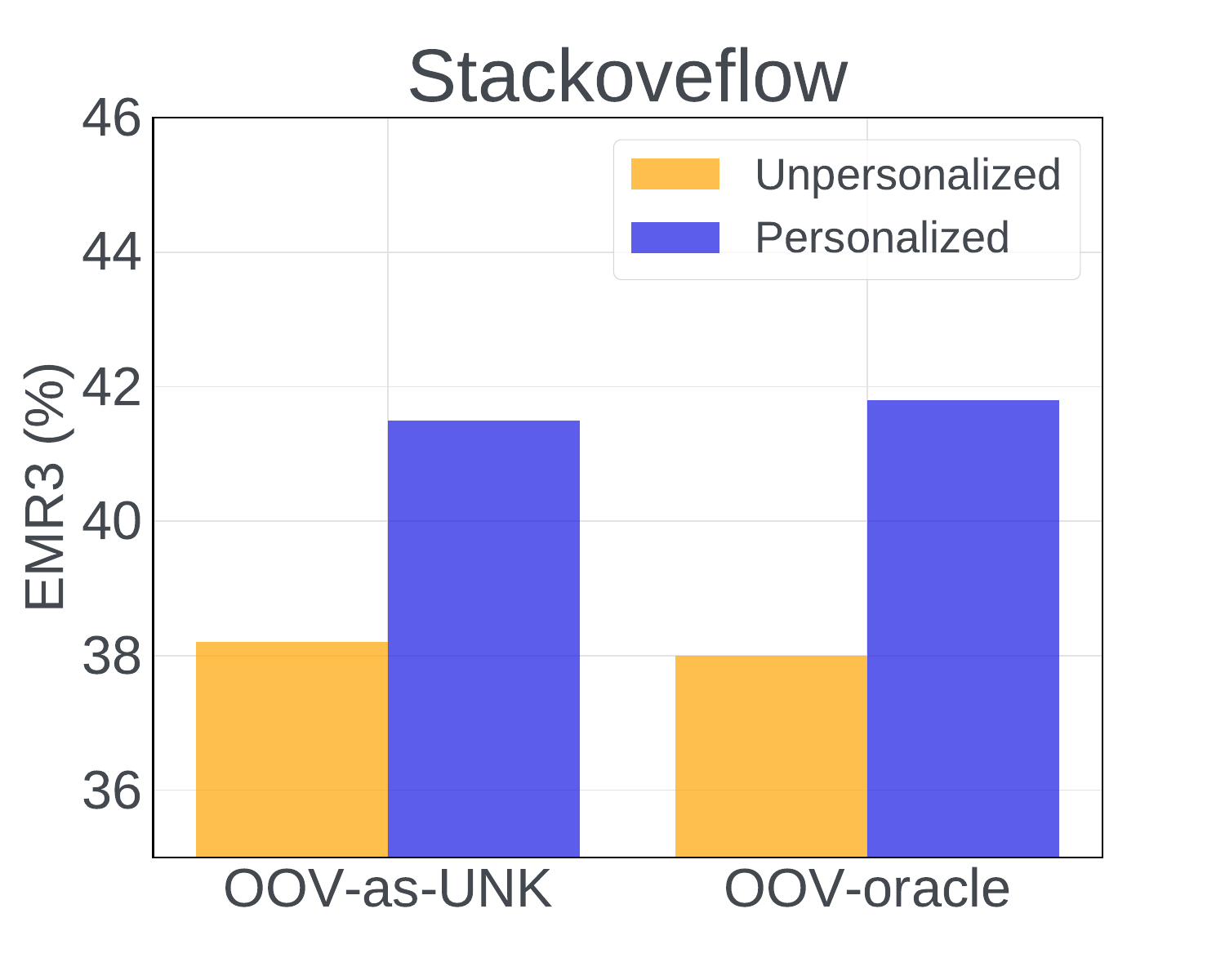}
     \end{minipage}
     \caption{$\text{EMR}_3$ of two baselines before and after personalizations on each dataset.}
     \label{p_fig}
\end{figure*}

\subsection{Datasets}
\label{fldata}

We train, validate, and test our approach on 2 publicly available benchmark datasets for federated learning: pushift.io’s Reddit \cite{Caldas2018LEAFAB}, Stack Overflow \cite{STACK}. Pushshift.io Reddit is a previously existing dataset extracted and obtained by a third party that contains preprocessed comments posted on the social network Reddit and hosted by pushift.io. Stack Overflow consists of questions and answers from Stack Overflow. We experiment the next-word-prediction task with federated learning, using the natural non-IID partitionings of all datasets. 

Each dataset admits a training/validation/test split on client IDs. The training clients are used for federated learning, validation clients for early-stopping/hyperparameter search, and test clients for metric report and personalization.

Each test client's dataset is further divided into training, validation, and test segments with ratio $8:1:1$, where we (1) personalize the global model locally on the training segment, (2) use the validation segment for early-stopping and hyperparameter search, and (3) report model performance on the test segments.

\subsection{Evaluation metric}
\label{evalmetric}

The model quality is measured by how often the actual next word appears among the top $K$ word predictions, denoted by $\text{EMR}_K$ (Exact Match Rate). More precisely, given a dataset $\mathcal{D}$ of sentences, 
\begin{equation}
\label{emr}
\text{EMR}_K(\mathcal{D}) = \frac{\sum_{S\in \mathcal{D}} \sum_{w \in S} \mathbbm{1}_{\{w\in \text{Top}_K\text{pred}(w)\}}}{\sum_{S\in \mathcal{D}} |S| }.
\end{equation}
This is a natural metric for real-time language model in production, where $K$ next-word suggestions are surfaced when a user is typing. In our experiments we set $K=3$.

The final metric is given below:
\begin{eqnarray*}
\begin{split}
\text{EMR}_3 & = \frac{\sum_{u\in\mathcal{U}}\sum_{S\in \mathcal{D}_u} \sum_{w \in S} \mathbbm{1}_{\{w\in \text{Top}_3\text{pred}(w)\}}}{\sum_{u\in\mathcal{U}}\sum_{S\in \mathcal{D}_u} |S| }\\
&= \frac{\sum_{u\in\mathcal{U}}\sum_{S\in \mathcal{D}_u} |S| \cdot \text{EMR}_3(\mathcal{D}_u)}{\sum_{u\in\mathcal{U}}\sum_{S\in \mathcal{D}_u} |S|}
\end{split}
\end{eqnarray*}
where $\mathcal{U}$ is the set of all test clients and $\mathcal{D}_u$ stands for the test segment (see Subsection \ref{fldata}) of client $u$'s dataset. Equivalently, the model quality is measured by $\text{EMR}_3$ on the centralized test segments of all test clients.

\begin{figure*}[t]
     \centering
     \begin{minipage}[t]{0.4\linewidth}
         \centering
         \includegraphics[width=\linewidth]{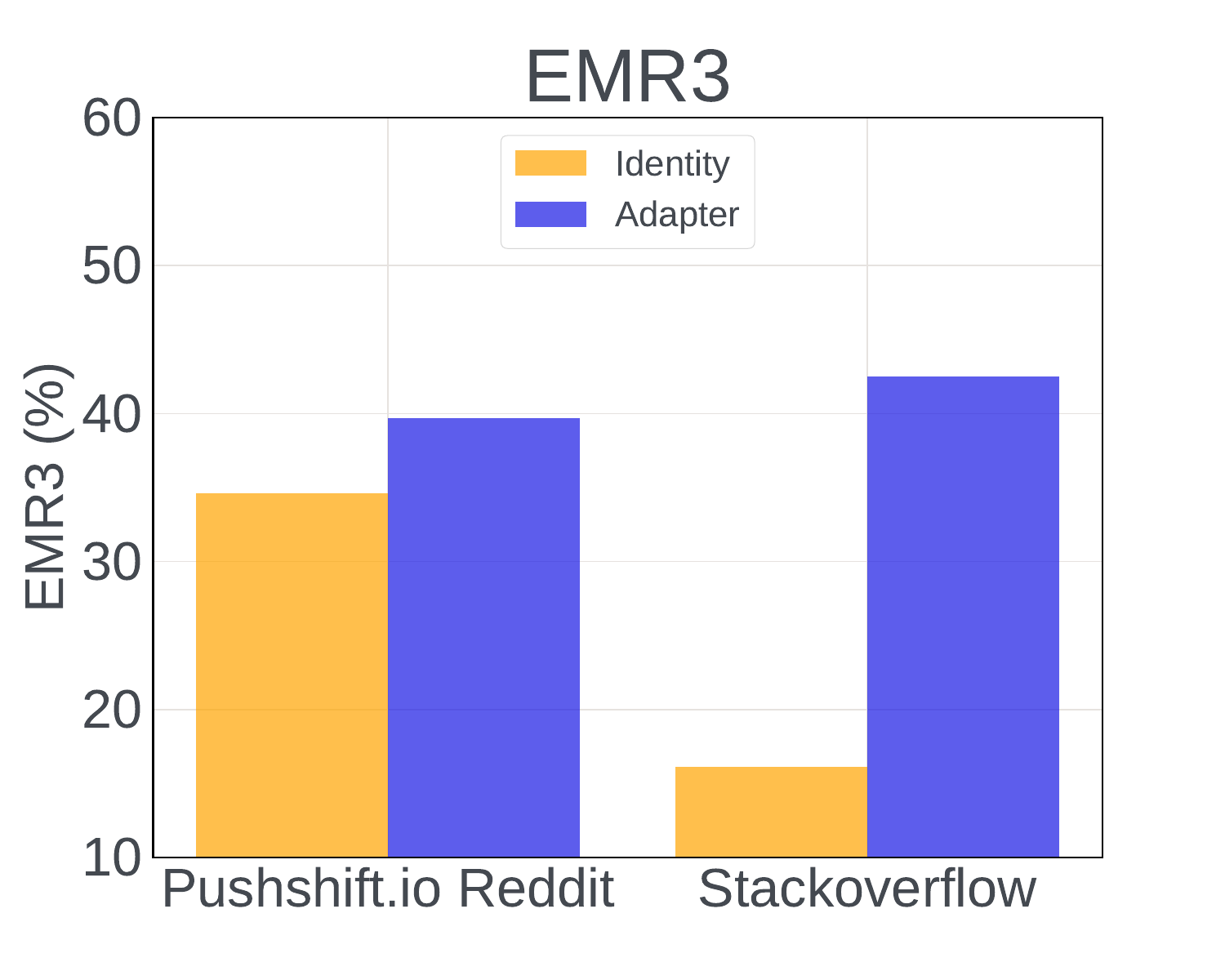}
     \end{minipage}
    \begin{minipage}[t]{0.4\linewidth}
         \centering
         \includegraphics[width=\linewidth]{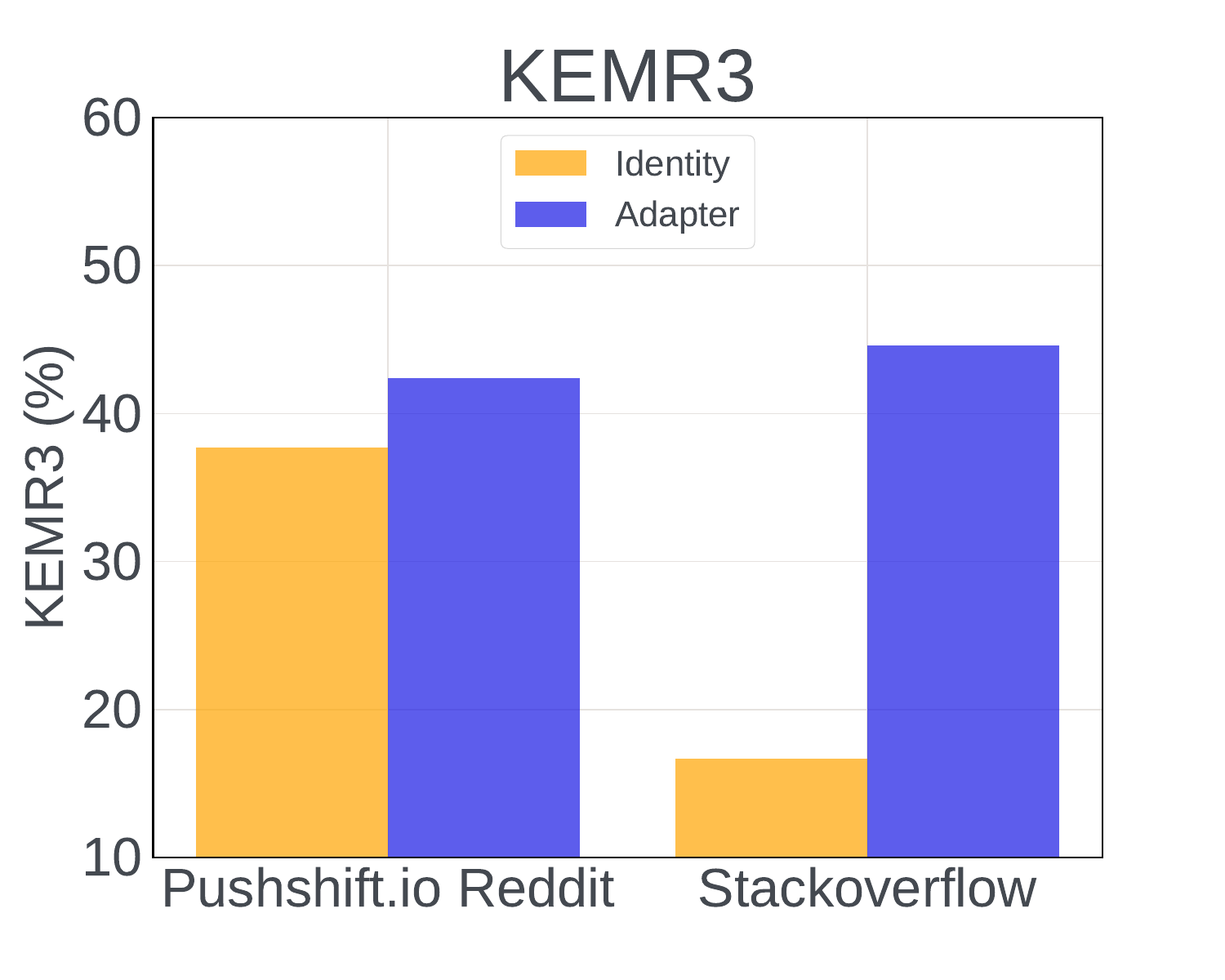}
     \end{minipage}
     \caption{$\text{EMR}_3$ and $\text{KEMR}_3$ of OOV expansion on 2 datasets, with and without adapter}
     \label{ablation_fig}
\end{figure*}

\subsection{Experimental details}
\label{fltraining}
The training recipes reported below are the same for all datasets unless stated otherwise. At each FL training round, 96 clients are randomly selected to participate in local training (evenly distributed to 8 GPU devices), with local epochs set to be 1 and training batch size as 8. Each global training epoch consists of $\#\text{users} / 96$ training rounds, where the total number of global training epochs are chosen to be 6 and 3 for Pushshift.io Reddit and Stackoverflow respectively.

We use SGD without momentum as client optimizer, and FedAdam for server side optimizer with weight decay of $10^{-5}$, $\beta_1$ as 0.9, $\beta_2$ as 0.999, and $\epsilon$ as $10^{-8}$. We do hyperparameter search on both client and server learning rates, with ranges $[10^{-6}, 0.05]$ and $[10^{-5}, 1]$ respectively. In Table \ref{besthps} we specify the best choices for both baselines on each dataset.

Every hyperparameter search contains 64 experiments (with 8 running in parallel at a time), where each single experiment takes 8 Nvidia V100 16GB GPU up to 10 hours and 12 hours for Pushshift.io Reddit and Stackoverflow respectively.

During personalizations, we perform hyperparameter search at each client among the following grid (if applied):
\begin{eqnarray*}
\begin{split}
\text{lr} &\in \{10^{-4}, 10^{-3}, 10^{-2}, 0.1, 1, 10\},\\
\sigma &\in \{0, 10^{-5}, 10^{-4}, 10^{-3}, 10^{-2}, 0.1, 1\},\\
\vec{\textbf{H}}&\in \{(960,), (128, 256, 128), (256, 512, 256)\}.
\end{split}
\end{eqnarray*}
Here lr denotes the learning rate for local training; $\sigma$ stands for the standard deviation of Gaussian distribution used to initialize the adapter parameters (except for LayerNorm); $\vec{\textbf{H}}$ represents the hidden dimensions of adapter. Note the grid only include $\sigma$ and $\vec{\textbf{H}}$ during OOV expansion. For each set of hyperparameters, we run 10 local epochs that early stops when the validation $\text{EMR}_3$ did not improve.

\begin{table*}[]\centering
\begin{tabular}{llll}
\hline
 & OOV-as-UNK & OOV-Oracle & OOV Expansion \\ \hline
OOV rate on Pushshift.io Reddit & 8.8\% & 5.5\% & 0.2\% \\ \hline
OOV rate on Stackoverflow & 4.8\% & 3.0\% & 0.001\% \\ \hline
\# Model parameters & 1.76M & 2.76M & 2.12M \\ \hline
\end{tabular}
\caption{Comparison of OOV rates and number of model parameters among 3 methods}
\label{oov_rate}
\end{table*}

\subsection{Results}
\textbf{Improves overall performance} As shown in Figure \ref{main_results}, our method achieves $2.1\%$ and $1.0\%$ (resp. $5.6\%$ and $2.5\%$) absolute (resp. relative) gains of $\text{EMR}_3$ compared to the standard approach (i.e. OOV-as-UNK) on Pushshift.io Reddit and Stackoverflow respectively. It shows competitive performance even when compared with the stronger baseline OOV-Oracle, seeing $3.7\%$ and $1.7\%$ relative $\text{EMR}_3$ gain on Pushshift.io Reddit and Stackoverflow respectively. The fact that OOV expansion yields better improvements on Pushshift.io Reddit relative to Stackoverflow partially attributes to the higher heterogeneity of Pushshift.io Reddit data, which is reflected by the larger OOV rates (see Table \ref{oov_rate}) and the worse long tail behavior (ref. Figure \ref{oov_tail}).\\ 
\\
\textbf{Increases word-coverage rates} As Table \ref{oov_rate} shows, the OOV expansion approach reduces the OOV rate of OOV-as-UNK (resp. OOV-oracle) by more than $97.7\%$ and $99.9\%$ (resp. $96.4\%$ and $99.9\%$) on Pushshift.io Reddit and Stackoverflow respectively.\\ 
\\
\textbf{More parameter efficient} As Table \ref{oov_rate} indicates, OOV expansion uses $24\%$ less parameters than the OOV-Oracle during personalization. In addition, our approach does not require any extra training parameters during pretraining or FL. Consequently, it is $36\%$ more parameter-efficient in FL compared to OOV-Oracle.\\ 
\\
\textbf{Personalization is essential} Taking Pushshift.io Reddit for instance, the standard personalization can already improve the global model accuracy by $6.5\%$ relatively on $\text{EMR}_3$ (see Figure \ref{p_fig}). With OOV expansion we can further boost this quality gain up to $12.5\%$.  \\ 
\\
\textbf{Adapter is necessary} As can be seen from the first plot in Figure \ref{ablation_fig}, adapter yields $15\%$ relative $\text{EMR}_3$ gain over trivial adapter (i.e. identity block) on Pushshift.io Reddit. For Stackoverflow dataset, inserting adapter achieves 2.6 times accuracy of the trivial adapter approach. In the second plot we present quality comparison under a new metric denoted by $\text{KEMR}_K$ which represents top $K$ known word exact match rates, and (similar to formula \eqref{emr}) it is defined by

\begin{table*}[]\centering
\begin{tabular}{lllll}
\hline
\multirow{2}{*}{} & \multicolumn{2}{l}{OOV-as-UNK} & \multicolumn{2}{l}{OOV-oracle} \\ \cline{2-5} 
 & client lr & server lr & client lr & server lr \\ \hline
Pushshift.io Reddit & 0.840 & 0.003 & 0.258 & 0.004 \\ \hline
Stackoverflow & 0.168  & 0.005 & 0.129 & 0.008 \\ \hline
\end{tabular}
\caption{Best learning rates for two baselines on 2 datasets}
\label{besthps}
\end{table*}

\begin{eqnarray*}
\begin{split}
\label{kemr}
\text{KEMR}_K(\mathcal{D}) = \frac{\sum_{S\in \mathcal{D}} \sum_{w \in S\cap \mathcal{V}} \mathbbm{1}_{\{w\in \text{Top}_K\text{pred}(w)\}}}{\sum_{S\in \mathcal{D}} |S\cap \mathcal{V}| }.
\end{split}
\end{eqnarray*}
where $\mathcal{V}$ is the closed vocabulary. 

From Figure \ref{ablation_fig}, we see that adapter not only helps achieves better OOV-understanding (i.e. higher EMR), the fact that it consistently improves accuracy on known word implies the important role of adapter in suppressing the issue of forgetting the knowledge from pretraining and federated learning stage.

\section{Related work}
Federated learning \cite{McMahan2016CommunicationEfficientLO} has achieved significant impact in the field of natural language processing (NLP) in recent years \cite{Chen2019FederatedLO,chenFL1,Wu2020FedMedAF,Lin2021FedNLPBF,Hilmkil2021ScalingFL,Liu2021FederatedLM,Ro2022ScalingLM}. Unlike non-federated language models where subword-based tokenizations such as WordPiece \cite{Schuster2012JapaneseAK}, Byte Pair Encoding (BPE) \cite{Sennrich2015NeuralMT} or SentencePiece \cite{Kudo2018SentencePieceAS} have become major choices, word-level tokenizer with a closed vocabulary is the default choice for edge device settings due to its low capacity and real-time nature. This gives rise to the commonly known OOV issue \cite{Chen2019GmailSC,chenFL1}, which has been investigated by several preceding work in federated learning: \cite{Chen2019FederatedLO} trains a separate character-level generative model to sample new words from, but by design requires an extra training stage; \cite{Singhal2021FederatedRP} introduces a new but more complex FL algorithm named FedRecon based on partial personalization; \cite{Bagdasaryan2022TrainingAT} proposes to iteratively update a subword-level tokenizer using the token sequences sampled from an FL-trained language model. Whereas \cite{knn_lm} and \cite{asr_ec} use memorization to dynamically update the vocabulary without retraining. However, these approaches either focuses on a different problem, or is not suitable for our settings due to their requirements of high training budgets, system complexity, and latency/memory costs. Instead, our approach is based on FL personalization, a technique that has driven a large body of work to address the presumed presence of heterogeneity \cite{Wang2019FederatedEO,Hanzely2020FederatedLO,Yu2020SalvagingFL,Fallah2020PersonalizedFL,Dinh2020PersonalizedFL,Li2020DittoFA,Singhal2021FederatedRP,Pillutla2022FederatedLW,Kulkarni2020SurveyOP}. Another key ingredient of our work is adapter \cite{Houlsby2019ParameterEfficientTL,Stickland2019BERTAP}, which has been widely studied in a large volume of work in the non-federated setting \cite{Mahabadi2021CompacterEL,Hu2021LoRALA,Li2021PrefixTuningOC,He2021TowardsAU,Pfeiffer2020AdapterFusionNT}.

\section*{Conclusion}
This paper proposes a personalized federated learning method that enables out-of-vocabulary words understanding for a class of on-device language models with closed vocabulary. By evaluating on two public benchmarks, we show that our method significantly outperforms the commonly used personalization approach in terms of next-word-prediction accuracy and drastically reduce the unknown-word rate on average while preserving user privacy and being parameter efficient.

\section*{Limitations}
One limitation of the results herein is the lack of study using subword-based language models. This was due to the tight memory, compute and latency budget of our experiments which made sub-word tokenization or beam search decoding infeasible. Another aspect that could be further explored is the trade off between accuracy and privacy by adding differential privacy to FL training. Last but not least, our method falls shy at cold-start problem (i.e. when a user's on-device historical data is insufficient or unavailable) because we assume to know all possible OOVs on-device.

\section*{Ethics Statement}
The proposed method in this work has the potential to improve the performance of language models for individual users by personalizing the vocabulary to their specific needs. This can have a positive impact in various applications including next word prediction, sentence completion and automatic speech recognition. Additionally, when used with federated learning, our approach can also help to protect the privacy of users by keeping personal data on individual devices and only sharing model updates with a centralized server. However, as is typical of any generative model, we also recognize that there are potential downsides to our approach. For example, if not implemented correctly, our approach could perpetuate existing biases in the data, specifically vocabulary and text usage patterns, and create unfair or biased models. Therefore, it is important for practitioners to be aware of these issues and take steps to mitigate them when implementing our method.

\bibliography{anthology,custom}
\bibliographystyle{acl_natbib}




\end{document}